\documentclass[preprint,review,12pt]{elsarticle}

\usepackage{amssymb}

\usepackage{url}

\usepackage{tabularx}
\usepackage{booktabs}
\usepackage{multirow}
\usepackage{rotating}
\usepackage{makecell}
\newcolumntype{P}[1]{>{\centering\arraybackslash}p{#1}}
\usepackage{subfigure}
\usepackage{xcolor}

\journal{Pattern Recognition}

\begin{document}

\begin{frontmatter}



\title{Deep Learning Techniques for Video Instance Segmentation: A Survey}


\author[label1]{Chenhao~Xu}
\ead{chenhao.xu@deakin.edu.au}
\author[label1]{Chang-Tsun~Li\corref{cor1}}
\ead{changtsun.li@deakin.edu.au}
\author[label2]{Yongjian~Hu}
\ead{eeyjhu@scut.edu.cn}
\author[label3]{Chee~Peng~Lim}
\ead{chee.lim@deakin.edu.au}
\author[label3]{Douglas~Creighton}
\ead{douglas.creighton@deakin.edu.au}

\cortext[cor1]{Corresponding author}
\affiliation[label1]{organization={School of Information Technology, Deakin University},
           city={Geelong},
           postcode={3216}, 
           state={VIC},
           country={Australia}}
\affiliation[label2]{organization={School of Electronic and Information Engineering, South China University of Technology},
           city={Guangzhou},
           postcode={510641},
           state={Guangdong},
           country={China}}
\affiliation[label3]{organization={Institute for Intelligent Systems Research and Innovation, Deakin University},
           city={Geelong},
           postcode={3216},
           state={VIC},
           country={Australia}}

\begin{abstract}
Video instance segmentation, also known as multi-object tracking and segmentation, is an emerging computer vision research area introduced in 2019, aiming at detecting, segmenting, and tracking instances in videos simultaneously. By tackling the video instance segmentation tasks through effective analysis and utilization of visual information in videos, a range of computer vision-enabled applications (e.g., human action recognition, medical image processing, autonomous vehicle navigation, surveillance, etc) can be implemented. As deep-learning techniques take a dominant role in various computer vision areas, a plethora of deep-learning-based video instance segmentation schemes have been proposed. This survey offers a multifaceted view of deep-learning schemes for video instance segmentation, covering various architectural paradigms, along with comparisons of functional performance, model complexity, and computational overheads. In addition to the common architectural designs, auxiliary techniques for improving the performance of deep-learning models for video instance segmentation are compiled and discussed. Finally, we discuss a range of major challenges and directions for further investigations to help advance this promising research field.
\end{abstract}



\end{frontmatter}


\section{Introduction}

Extending from image segmentation~\cite{wilson2003class, khadidos2017weighted}, Video Instance Segmentation (VIS) was initially proposed in~\cite{yang2019video} in 2019. In contrast to image segmentation, which only detects and segments objects in images, VIS involves more sophisticated and challenging instance tracking across video frames. 
VIS plays an important role in various real-world applications. 
As an example, by acquiring better representations of instances in videos, VIS assists in human action recognition and person (re-)identification, enhancing security for surveillance systems~\cite{algamdi2019learning, lin2020multi, lin2021detection}.
Given that Tesla is producing its DOJO supercomputer~\cite{talpes2023microarchitecture} to improve its driver-assistance system, VIS helps vehicles recognize and track other vehicles and pedestrians, boosting autonomous driving~\cite{alfasly2019auto, zhang2020traffic}. 
In the healthcare sector, VIS supports biomedical image analysis, pathology detection, and surgical automation~\cite{tan2019evolving, arbelle2022dual}. 
Furthermore, VIS demonstrates its potential to improve productivity, security, and user experience in the fields of agriculture~\cite{gan2022automated}, construction~\cite{xiao2022vision}, and entertainment~\cite{ghasemi2022deep}.

Deep learning is a machine learning methodology based on deep neural networks that consist of multiple layers and processing nodes~\cite{minaee2021image, xu2023scei}. Various deep neural networks, such as convolutional neural networks (CNN), recurrent neural networks (RNN), graph neural networks (GNN), and Transformers, have been increasingly adopted in deep-learning schemes for tackling challenges in the fields of computer vision~\cite{wang2022improving}, natural language processing~\cite{otter2020survey}, etc. These emerging deep-learning solutions usually demonstrate better performance than traditional machine-learning approaches~\cite{zhou2022survey}.

In recent years, numerous deep-learning schemes have been proposed for VIS. Typically, researchers propose novel deep-learning architectures by assembling mature deep neural networks, in order to more effectively extract features and aggregate spatiotemporal information. Besides, some researchers focused on auxiliary techniques, such as datasets and representation learning methodologies, to improve the performance of deep-learning models for VIS. In light of the rapidly expanding research attention on VIS, this paper reviews the existing works pertaining to deep-learning techniques for VIS. 

Numerous surveys on instance segmentation and object detection have been published in the literature. However, most of them focus on image segmentation~\cite{gu2022review}, Transformers~\cite{li2023transformer}, or video object tracking techniques~\cite{luo2021multiple, rakai2022data, javed2022visual}, with limited attention on the emerging VIS field. To close this gap, in this survey, deep-learning schemes for VIS are comprehensively reviewed, with key challenges and promising research directions identified. A comparison between this survey and existing survey papers is listed in Table~\ref{table:survey_comparison}. 

\begin{table}[htb]
\scriptsize
\caption{Comparison with Existing Survey Papers}
\label{table:survey_comparison}
\begin{tabularx}{\linewidth}{cc>{\raggedright\arraybackslash}p{0.3\linewidth}>{\raggedright\arraybackslash}p{0.5\linewidth}}
\toprule
\textbf{Ref} & \textbf{Year} & \textbf{Review Breadth} & \textbf{Review Depth}\\
\midrule
\cite{ciaparrone2020deep} & 2020 & Multi-Object Tracking & Deep Learning Techniques \\
\cite{yao2020video}       & 2020 & Video Object Segmentation and Tracking & Separate Segmentation and Tracking Methods \\
\cite{kalake2021analysis} & 2021 & Multi-Object Tracking & Real-Time Deep Learning Techniques \\
\cite{luo2021multiple}    & 2021 & Multi-Object Tracking & Similarity Computation and Re-identification Techniques \\
\cite{javed2022visual}    & 2022 & Multi-Object Tracking & Discriminative Filters and Siamese Networks \\
\cite{rakai2022data}      & 2022 & Multi-Object Tracking & Data Association Methods \\
\cite{zhou2022survey}     & 2022 & Video Object Segmentation \& Video Semantic Segmentation & Deep Learning Techniques \\
\cite{wang2022recent}     & 2022 & Multi-Object Tracking & Embedding Methods \\
\cite{bashar2022multiple} & 2022 & Multi-Object Tracking & Object Detection and Association Methods \\
\cite{gao2023deep}        & 2023 & Video Object Segmentation & Deep Learning Techniques \\
\cite{hou2023survey}      & 2023 & Moving Object Segmentation & Efficient Deep Learning Techniques \\
\cite{li2023transformer}  & 2023 & Visual Segmentation & Transformer-Based Methods \\
Ours & - & Video Instance Segmentation & Deep Learning Techniques \\
\bottomrule
\end{tabularx}
\end{table}

In summary, the contributions of this paper are as follows. Firstly, deep-learning techniques for VIS are reviewed and qualitatively compared from the architectural perspective. Secondly, auxiliary techniques that improve the performance of deep-learning models for VIS are outlined. Thirdly, a number of challenges and potential research directions are highlighted to promote further research in the field of VIS.

This paper is organized as follows. Section~\ref{sec:background} provides the background knowledge for readers to better understand related techniques. Section~\ref{sec:deep_learning_architectures} analyzes, compares, and summarizes different deep-learning schemes for VIS from the perspective of architecture, while Section~\ref{sec:miscellaneous_techniques} reviews auxiliary techniques used to enhance the performance of deep-learning models for VIS. Section~\ref{sec:challenges_directions} sheds light on several challenges and future research directions. Finally, Section~\ref{sec:conclusion} concludes this survey. The abbreviations used in this survey are summarized in Table~\ref{table:abbreviation}.

\begin{table}[htb]
\scriptsize
\caption{Abbreviation and Description}
\label{table:abbreviation}
\begin{tabularx}{\linewidth}{cl}
\toprule
\textbf{Abbreviation} & \textbf{Description} \\
\midrule
CNN & Convolutional Neural Network \\
FPN & Feature Pyramid Network \\
GNN & Graph Neural Network \\
LSTM & Long Short-Term Memory Network \\
MOT & Multi-Object Tracking \\
MOTS & Multi-Object Tracking and Segmentation \\
RNN & Recurrent Neural Network \\
RoI & Region of Interest \\
VIS & Video Instance Segmentation \\
ViT & Vision Transformer \\
VOS & Video Object Segmentation \\
VPS & Video Panoptic Segmentation \\
VSS & Video Semantic Segmentation \\
\bottomrule
\end{tabularx}
\end{table}

\section{Preliminaries}
\label{sec:background}

Before diving into analyzing recent studies in VIS, it is essential to explain the fundamental concepts relevant to this survey, including various video segmentation tasks and deep neural networks.

\subsection{Video Segmentation}

\begin{figure}[ht]
    \centering
    \subfigure[VOS]{
      \includegraphics[width=0.25\linewidth]{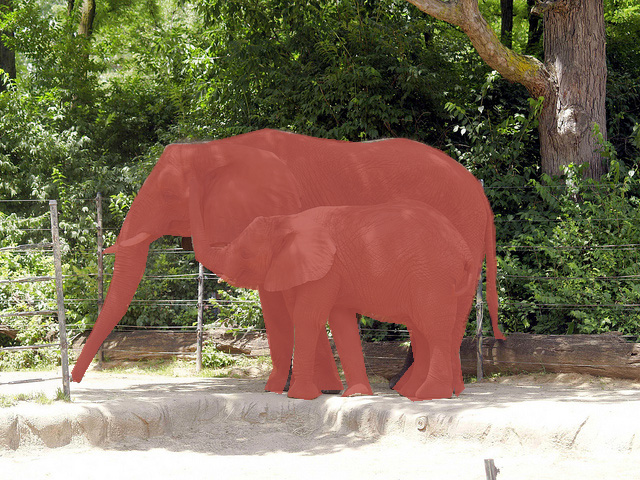}}
    \subfigure[VSS]{
      \includegraphics[width=0.25\linewidth]{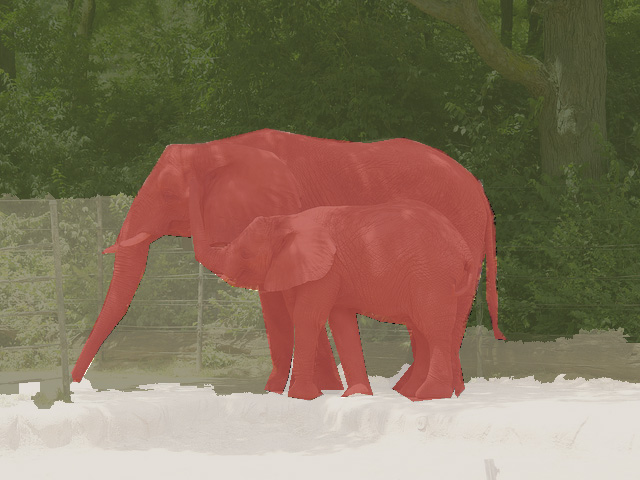}}
    \subfigure[VIS]{
      \includegraphics[width=0.25\linewidth]{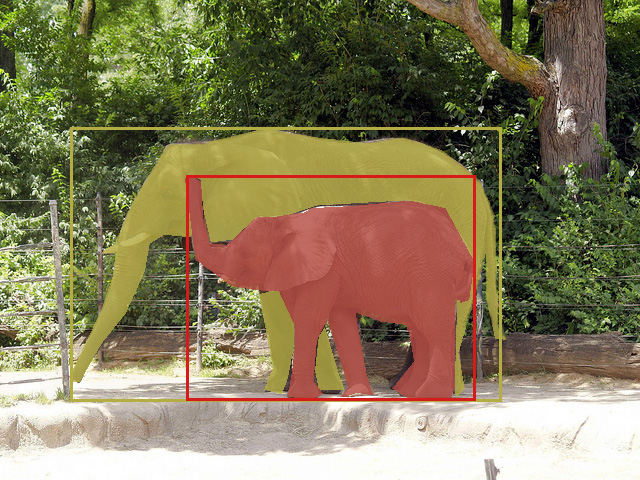}}
    \caption{Video segmentation tasks.}
    \label{fig:video_segmentation}
\end{figure}

Video segmentation aims to isolate and identify elements within a video. In particular, video segmentation encompasses several distinct tasks: video object segmentation, video semantic segmentation, and video instance segmentation, as shown in Fig.~\ref{fig:video_segmentation}. To help readers better grasp the extent of this survey, this section elaborates and compares these tasks. It is also important to note the difference between "objects" and "instances". An object in the context of VIS refers to a general category of items in a video frame (e.g., in a street scene, the objects could be pedestrians, cars, traffic signs, and buildings). An "instance" refers to an individual occurrence of an object category. For example, if there are multiple cars in a video frame, VIS would differentiate between each individual car by assigning it a unique label. Therefore, an object category encompasses multiple instances of that class of object.

\textbf{Video Object Segmentation}: Video Object Segmentation (VOS) is a binary segmentation task that requires the model to segment foreground objects from the background of a video~\cite{leyva2017video, yao2020video, zhou2022survey}. In other words, rather than segmenting every pixel in a frame, VOS only segments those pixels associated with the salient object. The classification results are binary, without separation of different instances of the same class of objects. VOS is the earliest video segmentation task and it serves as the basis for others.

\textbf{Video Semantic Segmentation}: Semantic segmentation was originally proposed for image processing, which requires the model to categorize each pixel in an image into a class~\cite{girshick2014rich, long2015fully}. Later, the concept of semantic segmentation was applied to videos, known as video semantic segmentation (VSS)~\cite{zhang2023semantic, slade2023neural}. Compared with VOS, VSS associates every pixel in a frame with one of multiple semantic categories. It is not necessary to discriminate different instances.

\textbf{Video Instance Segmentation}: In 2019, Yang et al.~\cite{yang2019video} introduced the VIS task, which requires the detection, segmentation, and tracking of individual instances of objects in videos. In the same year, Voigtlaender et al.~\cite{voigtlaender2019mots} extended the Multi-Object Tracking (MOT)~\cite{lin2021detection} to instance segmentation tracking and coined the term ``Multi-Object Tracking and Segmentation'' (MOTS), which is similar to VIS. The only difference between VIS and MOTS is that MOTS requires that masks do not overlap during evaluation~\cite{luiten2019video}. Therefore, in this survey, the two terms, VIS and MOTS, are used interchangeably. Compared with VOS and VSS, VIS segments salient objects in each frame and allocates the results into multiple classes, while identifying and tracking individual instances. 

\subsection{Deep Neural Networks}
As there are numerous deep neural networks, the most popular ones used in existing deep-learning schemes for VIS are introduced below.

\textbf{Convolutional Neural Network (CNN)}: A CNN is a popular deep neural network that automatically extracts features from images using its convolution kernels~\cite{xu2022efficient, xu2023scei, gu2018recent}. CNNs are widely applied in image and video regions, such as object detection, object tracking, action recognition, etc~\cite{gu2018recent}. 

\textbf{Recurrent Neural Network (RNN)}: An RNN is a deep neural network designed for sequential data or time series data, as it retains context information via cycles in the network~\cite{aafaq2019video}. As a result, several famous RNNs, such as the Long Short-Term Memory Network (LSTM), are widely adopted in VIS schemes to learn sequential visual features frame-by-frame with the help of CNNs~\cite{rafiq2023video}.

\textbf{Graph Neural Network (GNN)}: A GNN is a deep neural network designed for graph data, which captures dependency relationships among nodes via message passing between the nodes of graphs and conducts node, edge, and graph level predictions~\cite{wu2020comprehensive}. In VIS, the GNN is typically used to model the relationships among instances for better instance tracking and segmentation~\cite{wang2021end1}.

\textbf{Transformers}: A Transformer is a popular deep neural network with a self-attention mechanism that enables the global perception of a long sequence of tokenized inputs by automatically amplifying the key tokens~\cite{li2023transformer}. 
Vision Transformer (ViT) is a kind of Transformer that breaks down input images into a sequence of patches and then tokenizes them. Because of its outstanding performance, ViT is used for a growing number of computer vision tasks, such as image classification~\cite{dosovitskiy2020image}, object detection~\cite{carion2020end, sun2021sparse}, and VIS~\cite{he2022inspro}. Transformer is able to detect and segment objects at the frame level following the design of DEtection TRansformer (DETR)~\cite{carion2020end}. Transformer also offers long-range dependency modeling and temporal feature linkage for better instance tracking~\cite{wang2021end0}.

\textbf{Backbone, Neck, and Head}: To complete complex tasks in computer vision, deep-learning paradigms are typically composed of multiple kinds of deep neural networks organized as backbone, neck, and head~\cite{zhou2022survey, li2023transformer}. In particular, a backbone is responsible for extracting features from the input, a neck aggregates and refines the features extracted by the backbone, while a head is responsible for making predictions. These concepts are carried over into the deep-learning paradigms for VIS.

\section{Deep Learning Architectures for Video Instance Segmentation}
\label{sec:deep_learning_architectures}

In this section, the recent deep-learning schemes for VIS are analyzed and categorized from the perspective of architecture. In particular, as the backbone of the deep-learning schemes for VIS usually has a similar design for extracting frame-level features, the classification criteria mostly rely on the feature processing design in the neck. Specifically, deep-learning schemes for VIS can be broadly categorized into multi-stage, multi-branch, hybrid, integrated, and recurrent types. 
Table~\ref{table:architecture_comparison} outlines the pros and cons of different deep-learning architectures. Besides, Table~\ref{table:architecture_comparison} presents the design ideas for each deep-learning architecture and the corresponding works.

\begin{sidewaystable}
\scriptsize
\caption{Comparison of Deep-Learning Architectures for Video Instance Segmentation}
\label{table:architecture_comparison}
\begin{tabularx}{\textwidth}{p{0.07\linewidth}p{0.21\linewidth}p{0.2\linewidth}p{0.43\linewidth}}
\toprule
\textbf{Arch.} & \textbf{Design Ideas} & \textbf{Work} & \textbf{Pros ($\ast$) and Cons (-)} \\
\midrule
\multirow{5}{=}{M-Stage} & Mask R-CNN & 
\cite{yang2019video,voigtlaender2019mots,dong2019temporal,porzi2020learning,feng2019dual,luiten2020unovost,luiten2019video,choudhuri2021assignment} 
& $\ast$ Effective in extracting both low- and high-level features. \\
           & Mask Propagation     & 
\cite{tran2019guided,tran2020multi,bertasius2020classifying,lin2021video}        
& $\ast$ Easily replace sub-networks to suit various applications. \\
           & IRNet                & 
\cite{liu2021weakly,ruiz2021weakly}         
& - More processing stages increase computational complexity. \\
           & Attention Mechanism  &
\cite{liu2019spatio,fu2021compfeat,abrantes2023refinevis,cai2022dior,hu2021istr,zhang2023dvis}
& \\
           & Polamask \& FCOS     &
\cite{dong2021polarmask,liu2021sg}
& \\
\hline
\multirow{7}{=}{M-Branch} & Instance + Object Segment &
\cite{le2019semantic,lin2019agss,ge2021video,wang2023look}
& $\ast$ Effective for spatiotemporal feature processing. \\
           & Detection + Tracking &
\cite{wang2019empirical,liu2023instmove} 
& $\ast$ Effective for multi-modal feature processing. \\
           & YOLACT               &
\cite{bolya2019yolact,bolya2022yolactpp,bae2021occluded,cao2020sipmask,cao2022sipmaskv2,cao2022sipmaskv2,li2021spatial,liu2021yolactedge}
& - Increased architectural complexity \\
           & Siamese Network      &
\cite{wu2021track,yang2021crossover,wu2022defense,wu20221st,jiang2022stc,zhu2022instance,han2022visolo,li2022hybrid,li2022video0,wu20221st,wu2022defense,yan2022towards,yang2022less}
& - Requiring careful design and tuning to balance the branches. \\
           & Knowledge Distillation &
\cite{kim2023offline}
& \\
           & Point Clouds         &
\cite{xu2020segment,xu2021segment}
& \\
\hline
\multirow{4}{=}{Hybrid}     & \multirow{4}{=}{M-Branch Encoder \& Decoder}  &
\multirow{4}{=}{\cite{lin2020video,qin2021learning,qin2023coarse,zhou2021target,yan2022solve}}
& $\ast$ Better utilize the strengths of different types of networks. \\
&&& $\ast$ Effective for learning robust and generalized presentations. \\
&&& - Increased complexity and computational overheads. \\
&&& - Requiring careful selection and design of sub-networks. \\
\hline
\multirow{3}{=}{Integrated} & 3D-CNN \& GNN        &
\cite{athar2020stem,braso2022multi}
& $\ast$ Integrated feature processing for specific data distribution. \\
           & ViT                  &
\cite{cheng2021mask2former,choudhuri2023context,wang2021end0,hwang2021video,wu2022seqformer,zhang2023towards,yang2022temporally,zhang2023mobileinst,wu2022efficient,heo2022vita,huang2022minvis,ke2023mask}
& \makecell[tl]{- Requires a large dataset and long training for an ideal model. \\ - Not flexible enough to adjust for different purposes.} \\
\hline
\multirow{3}{=}{Recurrent}  & LSTM \& GNN          &
\cite{sun2019predicting,hu2021apanet,johnander2021video,brissman2023recurrent,wang2021end1}
& $\ast$ Effective for capturing temporal dependencies and context. \\
           & ViT with Query Propagation  &
\cite{meinhardt2022trackformer,koner2023instanceformer,heo2023generalized,you2022consistent,he2022inspro,ying2023ctvis,hannan2023gratt}
& - Longer contextual understanding entails more computational overheads. \\
\bottomrule
\end{tabularx}
\end{sidewaystable}

\subsection{Multi-Stage Feature Processing Architecture}

Multi-stage feature processing involves multiple feature processing and transformation stages in the neck, with each stage building upon the representations learned in the previous one. Earlier stages typically capture frame-level features and propose several Regions of Interests (RoIs), while later stages aggregate features and process abstract patterns and semantic information for tasks, such as object detection, object classification, instance segmentation, and instance tracking across frames. Popular multi-stage feature processing architectures for VIS include MaskTrack R-CNN~\cite{yang2019video} and TrackR-CNN~\cite{voigtlaender2019mots}, which are extended from a famous image instance segmentation network Mask R-CNN~\cite{he2017mask}.

When proposing the VIS task in~\cite{yang2019video}, Yang et al. extended Mask R-CNN~\cite{he2017mask} to MaskTrack R-CNN by adding a post-processing stage for tracking instances across video frames. Specifically, they utilize the memory queue to store the features of previously identified instances. A tracking head is embedded into Mask R-CNN to compare the similarity between the newly detected instance and the identified instances.
When introducing the MOTS task in~\cite{voigtlaender2019mots}, the authors proposed a TrackR-CNN network extended from Mask R-CNN~\cite{he2017mask}. In particular, the 3D-CNN is used for feature extraction~\cite{slade2023neural}. On top of the feature maps, the first stage utilizes a Region Proposal Network (RPN)~\cite{ren2015faster} to generate the proposal of target objects. In the second stage, several headers are used to predict the class, box, binary mask, and association vector for each RoI. The Euclidean distances between the association vectors are computed to associate the detected instances over time into tracks. 
As early works on VIS, MaskTrack R-CNN and TrackR-CNN~\cite{yang2019video, voigtlaender2019mots} simply extend the image instance segmentation architecture (Mask R-CNN), thus exhibiting several shortcomings, including segmentation precision, instance tracking consistency, occlusion resistance, and computational complexity. Most subsequent VIS schemes improve certain aspects and take these two schemes as benchmarks.

In terms of instance tracking, the authors of~\cite{dong2019temporal} utilized a CNN to extract features from multiple frames simultaneously and a Siamese network~\cite{lin2020multi} with cosine similarity to track the temporal features. Theoretically, this scheme avoids using the computationally expensive 3D-CNN, thus improving computing efficiency to some extent. However, the additional overheads for getting RoI proposals from multiple frames and the ensuing feature similarity comparison increase computational complexity. Besides, the randomly sampled frames from the video sequence cannot ensure the reliability of feature comparison. Similar to~\cite{dong2019temporal}, Porzi et al.~\cite{porzi2020learning} introduced a tracking head component to Mask R-CNN. The method accepts output from both the region segmentation head and the corresponding RoI features from the Feature Pyramid Network (FPN)~\cite{lin2017feature}. While the method eliminates the need for memory caching for existing instances, it also limits the capability of identifying re-appearing instances. To improve the performance of instance tracking and re-identification, the authors of~\cite{feng2019dual} proposed a bi-directional tracker, known as Instance-Pixel Dual-Tracker (IPDT). Based on the RoI proposals, the categories of objects are first calibrated to filter out false-positive classes within the global context of the video. Then, IPDT bidirectionally tracks instance-level and pixel-level embeddings, aiming at infusing the instance-level concept and discriminating overlapping instances. Instead of treating a video as an array of frames, in~\cite{luiten2020unovost}, the authors treat it as a tree composed of multiple tracklets in order to better track the re-appearing instances. In particular, a tracklet association algorithm, UnOVOST, was proposed based on Mask R-CNN~\cite{luiten2020unovost}. In the first stage, segments in consecutive frames are combined into short tracklets that contain segments from the same object by using spatiotemporal consistency cues. In the second stage, these short tracklets are merged into long-term tracks using decision trees, which are pruned based on the appearance similarity. Their approach improves the performance of long-distance instance tracking. However, the computational complexity and memory overhead are relatively higher when compared with those from frame-level tracking approaches. By adopting UnOVOST, in~\cite{luiten2019video}, the authors obtained first place in the 2019 YouTube-VIS challenge. Apart from decision trees, researchers have employed dynamic programming to identify global optimal assignments of instances across frames~\cite{choudhuri2021assignment}. A prominent advantage of these schemes is the improved understanding of instances across frames and better segmentation performance on overlaps, as compared with those from early VIS schemes MaskTrack R-CNN and TrackR-CNN.

A typical way to train a semi-supervised VIS model is to propagate masks of one or several keyframes to the entire video or video clips.
As an example, a bi-directional instance segmentation method was proposed by Tran et al. in~\cite{tran2019guided}. The work formulates a forward and backward propagation strategy to utilize masks in neighboring frames as references for instance segmentation at the current frame. The following year, Tran et al. introduced a multi-referenced guided instance segmentation scheme~\cite{tran2020multi}. After the first round of mask propagation, reliable frames are cached in memory as references for the second round of mask propagation. However, this two-pass processing approach is inefficient for dealing with long videos. Similarly, in~\cite{bertasius2020classifying}, Bertasius et al. proposed to propagate the instance features in a specific frame to the whole video clip by comparing the difference of the feature tensors, which enables a clip-level instance tracking. The approach performs better than MaskTrack R-CNN when handling overlapping instances. However, it appears to be challenging to separate dense instances in videos and identify fine-grained instance features~\cite{zhang2021refinemask}, as it relies too heavily on the mask prediction of Mask R-CNN. Inspired by the findings in~\cite{oh2019video}, the authors of~\cite{lin2021video} introduced a \textit{propose-reduce} paradigm for semi-supervised VIS. Specifically, based on Mask R-CNN, a sequence propagation head is appended to generate instance sequence proposals based on multiple keyframes in the video. Then, redundant proposals involving the same instances are reduced through various non-maximum suppression (NMS) techniques. This propose-reduce strategy is straightforward for tracking instances in long videos. Nonetheless, it is by no means trivial to strike a good balance between computing overhead and instance tracking performance when adjusting the number of sequence proposals. A more effective keyframe selection method is required to improve performance and computational efficiency.

Instead of mask propagation, another way to train a semi-supervised VIS model is to utilize the Inter-pixel Relation Network (IRNet)~\cite{ahn2019weakly}. IRNet is a CNN architecture built on a Class Attention Map (CAM). The CAM locates distinct instances and approximates their rough boundaries with only image-level supervision. In~\cite{liu2021weakly}, Liu et al. adopted IRNet with optical flow to assign similar labels to pixels with similar motion. Temporal consistency is leveraged to propagate trustworthy predictions between adjacent frames, in order to recover missing instances between frames. The method is superior to other image instance segmentation schemes, however, there is still a gap in precision when compared with supervised VIS schemes. Similarly, in~\cite{ruiz2021weakly}, Ruiz et al. proposed a weakly supervised learning strategy for MOTS based on gradient-weighted CAM (Grad-CAM)~\cite{selvaraju2017grad}. In essence, Grad-CAM is a class-discriminative localization technique that predicts a coarse localization heat map for the target concept. Since multi-task learning is incorporated in their scheme for predicting masks, bounding boxes, and classes simultaneously, Grad-CAM is utilized to locate the foreground mask based on the output of the classification branch.

Several studies leverage attention mechanisms to improve the detection, segmentation, and tracking capabilities of conventional multi-stage feature processing schemes. 
In~\cite{liu2019spatio}, Liu et al. embedded a spatiotemporal attention network to MaskTrack R-CNN. The aim is to focus on the instances belonging to pre-defined categories by estimating the attention on two consecutive frames.
In~\cite{fu2021compfeat}, Fu et al. introduced a frame-level attention module and an object-level attention module to Mask R-CNN for better RoI and object proposals. 
In~\cite{abrantes2023refinevis}, Abrantes et al. adopted the Transformer to conduct frame-level instance segmentation, followed by the temporal attention refinement of masks within tracklets.
Similarly, in~\cite{cai2022dior}, a temporal attention module and a spatial attention module were incorporated into Mask R-CNN to refine the instance-aware representations in videos.
A recurrent Transformer-based refinement strategy was used with Mask R-CNN to predict low-dimensional mask embeddings and improve performance~\cite{hu2021istr}.
However, such partially attention-improved schemes typically bring further complexity and computational overhead to classical multi-stage feature processing frameworks, with marginal improvement in performance. 
By abolishing the classical multi-stage feature-processing design, the work in~\cite{zhang2023dvis} decoupled the VIS schemes into three sub-tasks, i.e. segmentation, tracking, and refinement, with each of them handled by a different attention module. After segmenting instances at the frame level, a cross-attention mechanism~\cite{carion2020end} is leveraged to model the inter-frame association, thus tracking instances and achieving online VIS. Besides, an offline refiner module based on Transformer is devised for exploiting the context information from the entire video to refine the output of instance tracking.

In terms of efficiency, some researchers focus on reducing the computational complexity of models for VIS so that they can be applied to vehicles and other edge devices. Dong et al.~\cite{dong2021polarmask} introduced a lightweight VIS network, which regresses a group of fixed edge points in a polar coordinate system, rather than predicting conventional instance masks. After extracting features with the FPN, the centroid of instances is predicted based on the heat map, which is then utilized for predicting polar masks. The approach is capable of real-time tracking on mobile edge computing platforms. Another idea to improve computational efficiency is eliminating the predefined set of anchor boxes~\cite{tian2019fcos}. Liu et al.~\cite{liu2021sg} extended FCOS~\cite{tian2019fcos}, an anchor box-free and proposal-free object detection model, to VIS by incorporating an additional tracking head and a mask head. Target instances are dynamically divided into sub-regions based on their bounding box for fine-grained instance segmentation. The tracking head directly models object movements using object detection centers generated with FCOS to track instances. 
However, for these efficiency-focused methods, overcoming the occlusion and motion blur issues as well as improving segmentation precision remain challenging tasks.

\subsection{Multi-Branch Feature Processing Architecture}

The multi-branch feature processing architecture consists of multiple branches working in parallel to process different aspects or representations of the input data. The branches typically process features for different subtasks, and the outputs of these branches are often fused or combined to achieve VIS. Through multiple branches, the model can capture complementary information and learn robust and discriminative representations. In fact, multi-branch architectures typically can yield improved performance as compared to single-branch architectures, but at the expense of increased parameters and computational overhead during training and inference. Some representative studies utilizing this architecture are as follows.

The saliency map is a crucial cue for VIS to focus its search on salient regions in each frame~\cite{hossny2017enhanced, zhang2023semantic}. Standing on the viewpoint of two subtasks, semantic instance segmentation (SIS) and salient object segmentation (SOS), Le et al.~\cite{le2019semantic} proposed a Semantic Instance - Salient Object (SISO) framework with two branches in charge of these two subtasks, respectively. In terms of semantic instance segmentation, instance masks with high confidence at each frame are propagated and integrated into instances in later frames. In terms of salient object segmentation, a 3D fully convolutional network (3D-FCN)~\cite{le2017deeply} is adopted to compute salient region masks. By fusing the features from these two branches and integrating an identity-tracking module, semantic salient instances in the video are finally segmented. 
In~\cite{lin2019agss}, Lin et al. proposed a similar idea that simultaneously captures features shared by all instances (i.e., SOS) and discriminates different instances by instance-specific features (i.e. SIS). The framework consists of two branches, one of which is dedicated to an instance-agnostic module, and the other to an instance-specific module. The features from both branches are then fused by an attention-guided decoder, followed by a final prediction module. 
Similarly, Ge et al.~\cite{ge2021video} designed two branches based on the correlation matrix, with one generating coarse instance score maps and the other separating the foreground from the background. 
In~\cite{wang2023look}, the authors argued that instance understanding matters in VOS and developed a two-branch network. The instance segmentation branch explores the instance features of the current frame while the VOS branch performs spatiotemporal matching with the memory bank.
Despite yielding a good performance by adopting models pre-trained on public datasets, fine-tuning the models for both branches is more challenging than for multi-stage schemes. Besides, compared to single-branch approaches, multi-branch schemes typically have more parameters and lower processing efficiency.

Another method for solving VIS is assigning two branches to perform the detect and track operations, respectively. In~\cite{wang2019empirical}, a Hybrid Task Cascade (HTC)~\cite{chen2019hybrid} method was devised to conduct image instance segmentation, and SiamMask~\cite{wang2019fast} was utilized to track objects. Two original SiamMasks are cascaded to better predict the mask of instances. However, when dealing with overlapping instances, the performance is significantly influenced by the image instance segmentation module. 
To deal with situations where instances vanish and then re-appear, some researchers introduce a memory bank beside the instance mask prediction branch to store representative motion patterns~\cite{liu2023instmove}. Before being fed to a decoder to predict the target mask, the encoded mask features in the current frame are combined with the motion pattern representations retrieved from the memory bank, which improves robustness toward occlusion and fast-moving objects. However, the performance is subject to the motion patterns learned in previous frames. The extra computational and storage overheads caused by ConvLSTM and the memory bank are also non-negligible.

As the You Only Look Once (YOLO)~\cite{redmon2016you} algorithm becomes popular for object detection in images, an extended variant, named YOLACT~\cite{bolya2019yolact, bolya2022yolactpp}, demonstrated its effectiveness in VIS. The two-branch design of YOLACT for prototype mask generating and mask coefficient prediction enables efficient image-level instance segmentation. The effectiveness of YOLACT under severe occlusions was validated in~\cite{bae2021occluded}. 
Based on YOLACT, in~\cite{cao2020sipmask, cao2022sipmaskv2}, the authors proposed SipMask. The model makes greater preservation of the spatial information contained within an instance by dividing mask prediction into several sub-mask predictions. In addition to image-level instance segmentation tasks, SipMask is validated as effective for VIS tasks. Although it is an efficient solution for VIS, the scheme lacks the utilization of temporal information across frames in video. 
An idea to improve YOLACT was proposed by adding another tracking decoder branch to produce embedding vectors for all instances~\cite{chang2021yoltrack}. Denoted as YolTrack, it improves the inference speed to a real-time level, but at the expense of accuracy and precision. 
Moreover, in~\cite{li2021spatial}, the FPN features from two adjacent frames were fused to explore temporal correlations. The approach comprises two branches of YOLACT for frame-level instance segmentation and an additional temporal branch for fusion features between two consecutive frames. Nevertheless, the approach lacks the comprehension and tracking of instances in long videos, especially when dealing with instances that disappear and re-appear. 
On the other hand, to improve the efficiency of YOLACT for use on edge devices, the authors of~\cite{liu2021yolactedge} proposed YolactEdge. It reduces the computational overhead on non-keyframes of the video by computing only a subset of the features.

Multi-branch design of the Siamese network~\cite{lin2020multi} is naturally ideal for tracking instances across different frames by comparing their feature embeddings. 
To allow the tracking clues to better assist in detection, the work in~\cite{wu2021track} proposed an association module based on a Siamese network. It extracts re-identification embedding features in consecutive frames to improve segmentation in the current processing frame. 
In~\cite{yang2021crossover}, a novel crossover learning scheme was devised for VIS based on the Siamese network, named CrossVIS. In particular, crossover learning enables dynamic filters to learn background-irrelevant representations of the same instance at two different frames. 
Based on the Siamese network, contrastive learning appears to be an efficient way to learn representations. The work in~\cite{wu2022defense} made the embeddings of the same instance closer and the embeddings of different instances farther apart in the embedding space by comparing the adjacent frames. The method achieved an overall first place in the 2022 YouTube-VIS challenge~\cite{wu20221st}. 
Another contrastive learning-based VIS strategy proposed in~\cite{jiang2022stc} aimed to improve instance association accuracy. Inspired by~\cite{zhu2017bidirectional}, a bi-directional spatiotemporal learning scheme was introduced for training.
Although the Siamese network provides efficient feature comparison for instance tracking between consecutive frames in a video, it is nevertheless challenging for these schemes to understand a long video and re-identify those vanish and re-appearing instances.
To improve robustness against occlusions and reappearance, identification and association modules were leveraged to predict identification numbers and track instances~\cite{zhu2022instance}. The identification module detects new instances, assigns them identities, and encodes them into embeddings for further propagation across frames. With embeddings of the last frame stored, the association module effectively aids in propagating information from previous frames to the current frame. The approach obtains good performance for solving occluded VIS tasks.
Subsequently, in~\cite{han2022visolo}, the authors proposed a grid-structured VIS model, i.e., VISOLO, based on the image instance segmentation scheme, SOLO~\cite{wang2020solo, wang2020solov2}. Specifically, each frame in the video is divided into uniform grids. They serve as the basic unit for creating semantic category scores and instance masks. Then, the memory-matching module, which caches feature maps of previous frames, calculates the similarity measure between grids in different frames for instance tracking. The grid-level features, in comparison with convolutional features, are easier to reuse and share across numerous modules. This allows the method to preserve a longer history of feature maps and improve robustness against occlusions and reappearance.

As Transformers arise in VIS, there are various proposed schemes that use the Siamese network to build inter-frame attention between the target frame and the referring frame. 
In~\cite{li2022hybrid}, the authors introduced an intra-frame attention module with shared weight to the Siamese network for linking both instance- and pixel-level features in each frame. Besides, an inter-frame attention module is used to fuse hybrid temporal information and learn temporal consistency across frames. 
Similar attention-based temporal context fusion was adopted in studies reported in~\cite{li2022video0, wu20221st, wu2022defense, yan2022towards} for inter-frame instance association. 
Although inter-frame attention is useful for tracking instances across different frames, it is challenging to effectively select reference keyframes in long videos in a way that reduces computational complexity while improving the accuracy of instance tracking. 
As a result, an inter-clip attention scheme based on Transformers was proposed in~\cite{yang2022less}. Specifically, by comparing the similarity of features of both target and referring clips, the instance sequences in the target video are learned in a few-shot manner. However, only limited evidence on the effectiveness of the method was provided.

Knowledge distillation~\cite{wang2021knowledge} is a machine learning method that transfers knowledge from a large model (teacher) to a smaller one (student). It allows an online VIS model to learn a wealth of knowledge from an offline model for consistent instance tracking and segmentation. Kim et al.~\cite{kim2023offline} proposed an offline-to-online knowledge distillation (OOKD) for VIS. The devised query filtering and association (QFA) module filters out bad queries and links the instance features between offline and online models. By encoding object-centric features from a single frame and augmenting them with long-range global context distilled from the teacher model, the model demonstrates state-of-the-art feature matching and instance tracking capabilities. 

Point cloud is also an effective way to learn instance representation. In~\cite{xu2020segment, xu2021segment}, the authors built two 2D point clouds in two separate branches to learn features from the foreground and surrounding area. The features for segmentation, such as offset, color, category, and position, can be extracted from cloud points. However, the precision of the scheme highly relies on the number of points utilized, while the use of more points results in a significantly heavier computational burden.

\subsection{Hybrid Feature Processing Architecture}

Hybrid feature processing architectures integrate multi-stage and multi-branch architectures into an integrated framework. With multi-stage processing in each branch, the features are aggregated and processed at higher semantic levels, enabling better performance on each subtask. On the other hand, with multi-branch processing different subtasks of VIS, the features are learned in a robust and discriminative manner. However, hybrid architectures are typically more complex than multi-stage and multi-branch architectures, raising concerns among researchers on whether the improved performance compensates for the higher computational burden. The works discussed below are some examples of hybrid feature processing architectures characterized by a multi-branch encoder-decoder design.

Without initial masks in the first frame, the variational autoencoder was incorporated into Mask R-CNN to aid in capturing spatial and motion information shared by all instances~\cite{lin2020video}. Specifically, there is one encoder in the architecture for generating latent distribution, along with three parallel decoders assigned to three different branches. These branches are in charge of learning semantic information, providing attentive cues to reduce false negatives, and aggregating features from the encoder. However, the architecture almost doubles the complexity of the original Mask R-CNN model. Another similar work based on the encoder-decoder model is~\cite{qin2021learning, qin2023coarse}. Based on features from the image encoder, three decoders, spike decoder, position decoder, and appearance decoder, produce the latent distribution, offset vectors, and appearance embedding, respectively. Empirical results show that the approach outperforms MaskTrack R-CNN in AP by $3.5\%$ while being two times slower. 

Considering the high annotation cost for VIS, a two-stage network was proposed in~\cite{zhou2021target}. It includes a two-branch discrimination network (D-Net) for video object proposals and a two-branch target-aware tracking network (T-Net) for associating object proposals. In D-Net, one branch estimates the salient objects, whereas the other branch predicts the instance pixels. By comparing the object proposals in the current frame with historical tracking results, T-Net generates a segmentation score prediction for the target object. Although accuracy is slightly improved, the method places a large amount of computing overhead on both training and inference due to the lengthy and complex design.
On the other hand, a semi-supervised framework requiring only bounding-box labels was developed in~\cite{yan2022solve}. Optical flow in a branch is exploited to capture the temporal motion among instances, while depth estimation is used in another branch to provide spatial correlation between instances. A series of pseudo labels for salient instances are generated by leveraging the features from both optical flow and depth estimation branches. A bounding-box supervised puzzle solver further refines and assembles the sub-optimal masks and recovers the original instances. The method is comparable to fully supervised TrackR-CNN and MaskTrack R-CNN in performance. Nevertheless, there is still room for improvement in terms of long-video understanding and identifying re-appearing instances. Besides, it is promising to remove the reliance on bounding box labels during training.

\subsection{Integrated Feature Processing Architecture}

The integrated feature processing architecture typically extracts features of all frames in a video or clip together to build a 3D spatiotemporal feature volume. By aggregating the spatiotemporal features, the model, which typically consists of an encoder-decoder design, automatically learns the high-level representations of diverse instances across time and space, followed by a final prediction. Integrated architecture offers an elegant design when compared with multi-stage and multi-branch architectures, and gains popularity as the self-attention mechanism becomes widely used. However, it usually necessitates a lengthy training process, more training data, and more computational and memory resources. The works that utilize integrated feature processing architecture are reviewed, as follows.

In 2020, Athar et al.~\cite{athar2020stem} adopted an FPN to extract different scales of feature maps. The feature maps are then stacked along the temporal dimension for decoding using a 3D-CNN model. However, the performance of this method largely depends on the capability of the 3D-CNN model to build the 3D mask tube. Precisely segmenting the 3D mask tube requires large memory consumption for storing more fine-grained spatiotemporal features.
In order to extract background-irrelevant features and track instances throughout the entire scene, Bras{\'o} et al.~\cite{braso2022multi} constructed a graph on a set of frames with each node representing an object detection. Then, feature embeddings obtained by the CNN are propagated across the graph for several iterations using neural message passing~\cite{gilmer2017neural} for predicting instance masks for each RoI. Although the GNN is a promising solution for learning instance association and background-irrelevant features at the video level, its computational complexity is highly sensitive to the instance density in frames and the video length.

As the self-attention mechanism of Transformers helps direct attention to the features of target instances at the image level~\cite{carion2020end}, many attempts have been made to exploit this merit for 3D spatiotemporal feature extraction in VIS. 
In~\cite{cheng2021mask2former}, Cheng et al. extended a Transformer-based image instance segmentation model, i.e., Mask2Former~\cite{cheng2022masked}, to VIS. Masked attention is applied to the 3D spatiotemporal features for directly predicting a 3D mask for each instance across time. 
Choudhuri et al.~\cite{choudhuri2023context} demonstrated that employing absolute position encoding like~\cite{cheng2021mask2former} could cause object queries to heavily rely on the positions of the instances, causing an inability to recognize instance position changes. Thus, they proposed relative object queries on relative positional encoding for the Transformer to better capture an instance's position changes over frames.
In~\cite{wang2021end0}, the authors extended the Transformer-based image object detection model DETR~\cite{carion2020end} to VIS. Denoted as VisTR, the method consists of an instance sequence matching module to supervise the instance sequence across frames, as well as an instance sequence segmentation module to accumulate the mask features and predict the final mask sequences.
These Transformer-based schemes generate a sequence of instance predictions concurrently using all frames in a video or clip as the input, resulting in a substantial computational and memory overhead.
To reduce computational and storage overhead, Hwang et al.~\cite{hwang2021video} proposed a clip processing pipeline. It yields better performance over those per-frame methods and less memory usage over those per-video methods. Specifically, two Transformers are designed, one encodes each frame independently, and the other exchanges information between frames.

Note that the self-attention mechanism of Transformers typically involves explosive computations and memory overheads over the space-time inputs of the entire video, as it has quadratic complexity with respect to the input sequence~\cite{zhu2021deformable}. 
Deformable attention~\cite{zhu2021deformable} achieves smaller computational complexity than full attention, as it only pays attention to a small number of key sampling points around a reference point assigned to each query.
As a result, a multi-level deformable attention scheme, named SeqFormer~\cite{wu2022seqformer}, was designed to encompass both frame- and instance-level attention queries on videos. In particular, SeqFormer first performs independent frame-level box queries using deformable attention~\cite{zhu2021deformable}. Then, the instance query is conducted based on the features extracted by box queries on each frame, which generates the final segmentation mask sequence. 
Following SeqFormer~\cite{wu2022seqformer}, Zhang et al.~\cite{zhang2023towards} indicated the importance of multi-scale temporal information for VIS. They proposed TAFormer to incorporate both spatial and temporal multi-scale deformable attention modules in an encoder. While TAFormer performs marginally better than SeqFormer, it has more tuning parameters and computational complexity.
Apart from deformable attention, the MSG-Transformer~\cite{fang2022msg} is a computation-efficient variant of the self-attention mechanism in computer vision. Instead of applying full attention to images, MSG-Transformer adopts local attention to subregions, and introduces an additional messenger token to each subregion for exchanging information across different subregions. As a result, the work in~\cite{yang2022temporally} extended the MSG-Transformer to VIS to enable efficient computation, named TeViT. In particular, TeViT constructs patch tokens along with messenger tokens on all frames in the video, and shifts messenger tokens across the time dimension for capturing temporal contextual information. Compared with VisTR, TeViT achieves better video processing speed and instance segmentation precision.
Additionally, SeaFormer~\cite{wan2023seaformer}, a lightweight ViT with squeeze-enhanced axial attention, was leveraged to produce an efficient VIS scheme for mobile devices~\cite{zhang2023mobileinst}. 
To speed up the convergence of VisTR, EfficientVIS was proposed in~\cite{wu2022efficient} by leveraging the clip processing pipeline. In particular, EfficientVIS extends Sparse R-CNN~\cite{sun2021sparse} with self-attention to support clip-level queries and proposals. However, the performance of clip-level queries and proposals in EfficientVIS is highly subject to the precision of spatiotemporal RoI~\cite{he2017mask} on video clips.
On the other hand, considering that the dense spatiotemporal features extracted from videos are the key reasons for high computational complexity, VITA~\cite{heo2022vita} was formulated to extract only object-aware context through a frame-level object detector. By collecting the frame-level object tokens for the entire video, VITA builds the relationships between every detected object and achieves global video understanding. 
Moreover, a simple and computation-efficient Transformer-based VIS scheme, i.e., MinVIS~\cite{huang2022minvis}, was developed. MinVIS only trains a query-based image instance segmentation model. In the post-processing step, the instances are tracked by bipartite matching of query embeddings across frames. MinVIS also supports sub-sampling the annotated frames in training videos to further improve training efficiency.

In terms of annotation-efficient VIS, based on Mask2Former~\cite{cheng2021mask2former}, the work in~\cite{ke2023mask} introduced MaskFreeVIS to substitute the requirement for mask annotations with bounding box annotations during the training. Specifically, BoxInst~\cite{tian2021boxinst}, a bounding-box supervised image instance segmentation approach, is extended with the Temporal KNN-patch Loss (TK-Loss). The method identifies one-to-many matches across frames through an efficient patch-matching step, followed by a K-nearest neighbor selection. Empirical studies demonstrate that MaskFreeVIS outperforms certain fully-supervised models like EfficientVIS~\cite{wu2022efficient}.

\subsection{Recurrent Feature Processing Architecture}

The recurrent feature processing architecture involves recurrently extracting and processing features from frames along the temporal axes. By recurrently propagating the features of past frames to the current frame, the recurrent architecture design allows a model to track instances in videos with marginal memory overhead. In addition to RNNs, as ViT becomes prominent, some works also propagate the object queries in Transformers in this way, therefore they are included in this section. The following are some studies that utilize recurrent feature processing architectures.

The temporal dimension of videos allows features to be processed in a recurrent manner according to the temporal flow of frames. One recurrent model to process spatiotemporal features in video is ConvLSTM~\cite{shi2015convolutional, sun2019predicting}. It extends LSTM with convolutional structures to better capture spatiotemporal correlations. Specifically, Sun et al.~\cite{sun2019predicting} proposed a contextual pyramid ConvLSTMs to process multi-level spatiotemporal features extracted by the FPN, followed by a Mask R-CNN header~\cite{he2017mask} for predicting the instances in the next frame. The method has the benefit of being fast for real-time applications and for making fine-grained use of the features. 
There is another scheme, namely APANet, that improves ConvLSTM by adaptively aggregating spatiotemporal contextual information acquired at various scales to more accurately predict future frames~\cite{hu2021apanet}. The connections among each pair of ConvLSTM units are determined by neural architecture search (NAS). 
In a nutshell, these ConvLSTM-based schemes require significant memory due to many spatiotemporal features being cached, causing them to struggle to comprehend long videos.

In addition to ConvLSTM, several researchers also employed a GNN along with LSTM to propagate information for tracking~\cite{johnander2021video, brissman2023recurrent}. In particular, a graph is built on the past and current detected instances. It is then utilized to produce output embeddings for association. The embeddings are then fed to LSTM for historical information aggregating and future tracking. It is obvious that GNN aids in building better associations of instances across frames. However, the approach heavily depends on the accuracy of instance detection. A false or missed detection in specific frames may have a huge impact on the scoring of instance connections, thus affecting the tracking and segmentation of instances in a video. A similar idea of adopting GNN for VIS was proposed in~\cite{wang2021end1}. Two consecutive frames, a reference frame and a target frame, are utilized to construct a graph and obtain aggregated spatiotemporal features. Without the help of ConvLSTM or LSTM, the method caches the history mask information in memory and achieves a similar effect of mask propagation. 

Recently, the self-attention mechanism is being increasingly utilized to construct query-based VIS schemes~\cite{fang2021instances}, where query proposals are typically propagated across frames for tracking instances~\cite{choudhuri2023context}. 
Meinhardt et al.~\cite{meinhardt2022trackformer} proposed a query-based VIS scheme, i.e., TrackFormer, based on Deformable DETR~\cite{zhu2021deformable}. It enables the transformer to detect and track objects in videos in a frame-by-frame manner. The tracking-by-attention paradigm and the concept of auto-regressive track queries are defined.
Similarly, Koner et al.~\cite{koner2023instanceformer} proposed a Transformer-based online VIS framework, denoted InstanceFormer. It incorporates a memory queue to propagate the representation, location, and semantic information of prior instances to achieve better instance tracking consistency.
Considering the aforementioned methods only handle inter-frame associations, Heo et al.~\cite{heo2023generalized} argued that the main bottleneck in processing long videos is building inter-clip associations. As a result, they proposed a clip-level query propagation approach, i.e., GenVIS, based on VITA~\cite{heo2022vita}. In particular, GenVIS stores clip-level decoded object queries in the memory. With the joint effort of decoded object queries propagated from the latest clip, GenVIS achieves state-of-the-art performance in long-distance instance tracking with a small computational overhead. 

To enhance temporal consistency in query propagation across frames, the work in~\cite{you2022consistent} employed additional clip-level queries for fusing information from all the frames. The scheme combines the designs of recurrent and integrated instance queries, thus improving the temporal consistency and robustness of query propagation on VIS tasks. However, the design increases computational complexity and memory overhead, and sacrifices the ability of real-time inference. 
Apart from introducing additional clip-level queries, caching instance features from previous frames is also helpful in improving temporal consistency. 
The work in~\cite{he2022inspro} introduced InsPro, which propagates query-proposal pairs from the previous frame to the current frame based on a set of instance queries. By caching instance features in historical frames and calculating intra-query attention, the method takes advantage of temporal clues in videos and copes well with occlusion and motion blur. In~\cite{ying2023ctvis}, the authors constructed contrastive items and added noise to the relevant embeddings in the memory bank during training to simulate identity switching in real-world scenarios, in order to better associate instances across time.
Another way to improve the temporal consistency and robustness of query propagation is to directly rectify the effect of noisy features accumulated during occlusion and abrupt changes. Hannan et al.~\cite{hannan2023gratt} proposed Gated Residual Attention for VIS (GRAtt-VIS), which uses gate activation as a mask for self-attention. The mask restricts the unrepresentative instance queries in the self-attention and keeps crucial information for long-term tracking. Compared with~\cite{you2022consistent}, the approach reduces computational complexity, alleviates memory overhead, and supports online processing. However, the shortcoming appears in tracking identities pertaining to crossover trajectories.

\section{Auxiliary Techniques for Enhancing Video Instance Segmentation}
\label{sec:miscellaneous_techniques}

In addition to the aforementioned architecture designs, there are several auxiliary techniques that can improve the performance of VIS, such as new datasets and representation learning techniques.

\textbf{Datasets}: Despite the fact that there are numerous datasets for instance segmentation, object detection, and semantic segmentation, most of them are prepared at the image level and only a few are specifically made for VIS. Table~\ref{table:datasets} summarizes the primary datasets for VIS that feature videos with multiple categories and with distinct instances annotated. In particular, YouTube-VIS~\cite{yang2019video} is the first large-scale and the most extensively adopted dataset for VIS, which is now updated to its 2022 edition. In~\cite{ke2022video}, the authors refined the masks in YouTube-VIS to High-Quality YTVIS (HQ-YTVIS). NuImages~\cite{caesar2020nuscenes} is distinguished by its attribute annotations, such as whether a motorcycle has a rider, the pose of a pedestrian, and the activity of a vehicle. OVIS~\cite{qi2022occluded} is a large-scale VIS dataset with a high percentage of occluded instances, which poses great challenges for VIS models.

\begin{table}[htb]
\scriptsize
\caption{Datasets for Video Instance Segmentation}
\label{table:datasets}
\begin{tabularx}{\textwidth}{p{0.18\linewidth}cccccl}
\toprule
\textbf{Dataset} & \textbf{Year}\textsuperscript{1} & \textbf{\#Video} & \textbf{\#Class} & \textbf{\#Mask} & \textbf{Scenario} & \textbf{Highlight} \\
\midrule
KITTI MOTS~\cite{voigtlaender2019mots} & 2019 & 21 & 2   & 38k    & Driving &  \\
SESIV~\cite{le2019semantic}        & 2019 & 84     & 29  & 12k    & General &  \\
BDD100K MOTS~\cite{yu2020bdd100k}  & 2020 & 90     & 10  & 129k   & Driving &  \\
NuImages~\cite{caesar2020nuscenes} & 2020 & 1,000  & 23  & 800k   & Driving & Attribute annotations \\
UVO~\cite{wang2021unidentified}    & 2021 & 11,361 & -   & 1,676k & General & Open-world mask \\
YouTube-VIS~\cite{yang2019video}   & 2022 & 4,019  & 40  & 266k   & General &  \\
OVIS~\cite{qi2022occluded}         & 2022 & 901    & 25  & 296k   & General & Heavy occlusion \\
VIPSeg~\cite{miao2022large}        & 2022 & 3,536  & 124 & 926k   & General & VPS \\
HQ-YTVIS~\cite{ke2022video}        & 2022 & 2,238  & 40  & 131k   & General & Fine-grained mask \\
BURST~\cite{athar2023burst}        & 2023 & 2,914  & 482 & 600k   & General &  \\
\bottomrule
\end{tabularx}
\noindent{\scriptsize{\textsuperscript{1} The release year of the latest version.}}
\end{table}

\textbf{Representation Learning}: In VIS, representation learning is a technique that helps VIS schemes to better extract features, capture motion patterns, reduce data requirements, and improve robustness and generalization. Several related works in this area are summarized as follows. 

As the FPN has been increasingly adopted in various VIS schemes, the authors of~\cite{li2022improving} proposed a Temporal Pyramid Routing (TPR) strategy that learns temporal and multi-scale representations altogether. Specifically, TPR accepts two feature pyramids from two adjacent frames as inputs. A Dynamic Aligned Cell Routing strategy is designed for aligning and gating the pyramid features across the temporal dimension. A Cross Pyramid Routing strategy is also proposed for transferring temporally aggregated features across the scale dimension. By incorporating the features from multiple frames, these representation-learning techniques improve clip-level instance understanding. However, they also impose additional memory overhead and computational complexity.

To learn high-quality embedded features, the connection between the instance segmenter and the tracker has been investigated. In particular, to increase randomness in training and encourage the tracker to learn more discriminative features, a sparse training and dense testing strategy was developed in~\cite{gao2022object}. The number of points sampled for training is fewer than that for testing. Additionally, a time-series sampling strategy that samples at random intervals ensures effective learning of temporal information. The approach not only facilitates the learning of more generalized and robust representations, but also reduces memory consumption during the training.

To fully exploit the pixel-wise annotations and increase the number of instances during the training, a data augmentation strategy, named continuous copy-paste (CCP), was proposed for VIS~\cite{xu2021continuous}. In particular, CCP retrieves several instance blocks from near frames and past them onto their original positions while mimicking their emerging and leaving by shifting two of them to the boundary. By preserving the relative offset of crops and the original positions of instances without modeling the surrounding visual context, CCP produces high-quality triplets for tracking. 
On the other hand, Yoon and Choi~\cite{yoon2023exploring} believed that models trained from representative frames with less redundancy could achieve comparable performance to that trained from dense datasets, thus reducing the cost of data acquisition and annotation. Specifically, an adaptive frame sampling (AFS) scheme is devised for extracting keyframes based on the visual or semantic dissimilarity between consecutive frames. With a simple copy-paste data augmentation on the keyframes, the performance gap caused by frame reduction is bridged.

\section{Challenges and Future Research Directions}
\label{sec:challenges_directions}

Although substantial progress has been made in VIS in recent years, there still remain numerous challenges. This section uncovers these challenges and proposes directions for future research and innovations in VIS.

\textbf{Occluded Video Instance Segmentation}: It is a challenge to segment highly-occluded instances in videos~\cite{wei2014dynamic}. The advent of the OVIS~\cite{qi2022occluded} dataset paves the way for further study in this area. In particular, the authors of OVIS defined a metric named Bounding-box Occlusion Rate (BOR) to reflect the degree of occlusion between objects, showing that OVIS has three times higher occlusions than the popular YouTube-VIS dataset. Based on OVIS, Ke et al.~\cite{ke2021deep, ke2023occlusion} addressed occlusion by treating each frame as a composition of two overlapping layers. In particular, a bilayer convolutional network is devised, which feeds the RoI features~\cite{tian2019fcos} into two branches for segmenting occluding objects (occluders) and partially occluded instances (occludees), respectively. In contrast to other amodal methods, which regress single occluded object boundary directly on the single-layered image, this approach takes into account interactions between the occluder and occludee. Nonetheless, there is still room for performance improvement by further utilizing contextual information propagated from adjacent frames. 

\textbf{Motion-Blurred Video Instance Segmentation}: Motion blur refers to the appearance of objects in a frame as being smeared or distorted due to a moving subject or camera, which usually occurs in sports videos and can adversely affect the performance of VIS~\cite{leyva2019compact, li2021spatial}. Since no datasets have been created specifically for this challenge, data augmentation can be exploited to synthesize the appearance of motion blur, and a metric to assess the degree of motion blur is also required. To precisely segment motion-blurred instances in videos, several directions of research are necessary, such as deblurring, motion estimation, blur-invariant feature extraction, and multimodal feature fusion. In~\cite{li2021spatial}, the authors fused temporal features from two adjacent frames to estimate the motion directions for better tracking instances in motion-blurred videos. While the method is useful, a systematic assessment and analysis of the VIS performance in terms of motion blur is necessary.

\textbf{Annotation-Efficient Video Instance Segmentation}: Given the high annotation cost for videos, it is encouraging to develop annotation-efficient VIS schemes, such as self-supervised~\cite{lin2023self}, weakly-supervised, or unsupervised VIS schemes~\cite{pourpanah2022review}. Caron et al.~\cite{caron2021emerging} demonstrated that self-supervised ViT features contain explicit information pertaining to the semantic segmentation of an image. Without using any labels, their proposed knowledge distillation approach, denoted as DINO~\cite{caron2021emerging}, automatically learns class-specific features in images by predicting the output of a teacher network using a cross-entropy loss. Based on DINO~\cite{caron2021emerging}, the work in~\cite{wang2023cut} proposed an unsupervised image segmentation scheme, i.e., CutLER, and applied it to VIS. CutLER outperforms other unsupervised VIS schemes significantly.
However, there are still gaps between annotation-efficient VIS schemes and fully supervised VIS schemes in terms of performance, prompting researchers to further exploit the available information in videos and make better use of weak annotations.

\textbf{Video Panoptic Segmentation}: In 2020, Kim et al.~\cite{kim2020video} introduced the term ``Video Panoptic Segmentation'' (VPS) as image panoptic segmentation began to gain popularity. In addition to the requirements in VIS, VPS demands models to segment every pixel in frames, including background elements. Although several VPS schemes have been proposed~\cite{kim2020video, qiao2021vip, miao2022large, li2022video1}, there is still room for improvement in prediction accuracy, segmentation refinement, training and inference efficiency, dataset diversity, and annotation efficiency. Particularly, in 2023, Athar et al. proposed a unified scheme for multiple video segmentation tasks, including VOS, VIS, and VPS~\cite{athar2023tarvis}. By modeling the targets of various tasks as different abstract queries of a Transformer, the method offers a viable path for a unified video segmentation solution and narrows the gap between VPS and VIS. 

\textbf{Open-Vocabulary Video Instance Segmentation}: Open-vocabulary VIS is a novel video segmentation task that requires the model to detect, segment, and track instances from open-set vocabulary categories, including novel categories unseen during training~\cite{thawakar2023video, wang2023towards}. Open-vocabulary VIS is highly valuable in real-world applications, especially when the object vocabulary is not fixed, such as surveillance and autonomous driving. In~\cite{wang2023towards}, Wang et al. proposed a Large-Vocabulary Video Instance Segmentation (LV-VIS) dataset along with a benchmark approach. The work paves the way for further research in this direction. Despite the fact that several early Transformer-based schemes have been proposed~\cite{thawakar2023video, wang2023towards, guo2023openvis}, the performance of Open-Vocabulary VIS lags behind that of classical VIS, owing to challenges in object diversity, data annotation, and semantic understanding. Several research directions, including zero-shot learning, adaptive learning, and multimodal learning, have great potential for developing more general Open-Vocabulary VIS models.

\textbf{Multimodal Video Instance Segmentation}: Multimodal VIS requires models to fuse features from various modalities and utilize their complementary properties~\cite{jaafar2023multimodal}. As the Transformer is effective in modeling global and long-range dependencies across different tokens~\cite{xu2023multimodal}, some researchers have utilized the Transformer to build multimodal VIS schemes. Botach et al.~\cite{botach2022end} and Chen et al.~\cite{chen2023vlkp} investigated the fusion of video and language features, while Li et al.~\cite{li2022online} focused on the fusion of video and audio features. Nevertheless, multimodal VIS still faces multiple challenges, such as multimodal data fusion and alignment, diverse data representation handling, and cross-modal data annotation collection. Incorporating generative models, like Make-A-Video~\cite{singer2022make}, which generates temporally coherent video clips from text, has the potential to mitigate the data-hungry issue of multimodal VIS.

\textbf{Promptable Video Segmentation}: In 2023, Kirillov et al.~\cite{kirillov2023segment} proposed a promptable segmentation task for images, which requires a model to accept flexible prompting (points, boxes, text, and masks) and return a valid segmentation mask in real time. With an innovative data engine for promptable segmentation, an incredibly huge and diverse set of masks has been created to train a Segment Anything Model (SAM). SAM enables zero-shot generalization, addressing novel visual concepts while resolving a variety of downstream segmentation issues. With the great success of promptable segmentation in image, promptable video segmentation holds promise for providing uniform solutions to various video segmentation tasks. However, compared with promptable segmentation in images, it is challenging to design video prompts. This is because it is difficult for mouse-driven points to follow an instance in a video consistently, which can easily lead to ambiguity. Besides, promptable video segmentation necessitates additional real-time tracking, prediction, and re-identification for instances across frames, posing challenges to real-time video understanding.



\section{Conclusion}
\label{sec:conclusion}

VIS is a fundamental computer vision task with extensive applications in numerous domains. VIS has made significant progress over the years, in line with the rapid development of deep-learning techniques and rising computing power around the world. To help researchers better understand the methodologies in this emerging field, this survey systematically reviews, analyzes, and compares existing deep-learning schemes from the perspective of architecture. Specifically, the reviewed schemes are divided into multi-stage, multi-branch, hybrid, integrated, and recurrent varieties according to their feature processing modes. Several auxiliary techniques for improving VIS performance, including specialized datasets and representation learning approaches, are scrutinized and discussed, providing readers with comprehensive research views on VIS. This survey also reveals several promising research directions by examining the key challenges faced by VIS, offering researchers with valuable insights into the advancement of video segmentation.

\bibliographystyle{elsarticle-num} 
\bibliography{ref}

\end{document}